\pdfoutput=1

\documentclass[11pt]{article}

\usepackage{ACL2023}

\usepackage{times}
\usepackage{latexsym}

\usepackage[T1]{fontenc}

\usepackage[utf8]{inputenc}

\usepackage{microtype}

\usepackage{inconsolata}

\usepackage{amsmath,color}
\usepackage{amsfonts}
\usepackage{amssymb}
\usepackage{amsthm}
\usepackage{mathtools}
\usepackage{enumerate}
\usepackage{graphicx,rotating}
\usepackage{subcaption}
\usepackage{float}
\usepackage{booktabs}

\usepackage{algorithm}
\usepackage{algpseudocode}

\theoremstyle{definition}
\newtheorem{definition}{Definition}

\theoremstyle{plain}

\newtheorem{theorem}{Theorem}

\theoremstyle{remark}

\newcommand{\eat}[1]{\ignorespaces}

\usepackage{chngcntr}

\title{Pareto Optimal Learning for Estimating Large Language Model Errors}

\author{
  Theodore Zhao \\
  \texttt{theodorezhao@microsoft.com} \And
  Mu Wei \\
  \texttt{mu.wei@microsoft.com} \\\AND
  Joseph S. Preston \\
  \texttt{sam.preston@microsoft.com} \\ \And
  Hoifung Poon \\
  \texttt{hoifung.poon@microsoft.com}
}

\begin{document}
\maketitle
\begin{abstract}
Large Language Models (LLMs) have shown impressive abilities in many applications. When a concrete and precise answer is desired, it is important to have a quantitative estimation of the potential error rate. However, this can be challenging due to the text-in-text-out nature of generative models. We present a method based on Pareto optimization that generates a risk score to estimate the probability of error in an LLM response by integrating multiple sources of information. We prove theoretically that the error estimator optimized in our framework aligns with the LLM and the information sources in an Pareto optimal manner. Experimental results show that the risk scores estimated by our method are well correlated with the true LLM error rate, thus facilitating error correction. By dynamically combining with prompting strategies such as self-verification and information retrieval, we demonstrate the proposed method can be utilized to increase the performance of an LLM, surpassing state-of-the-art task specific models.

\end{abstract}

\section{Introduction}
\label{sect-intro}

Large Language Models (LLMs) have evolved to become impressively powerful in recent developments \citep{zhao2023survey}, with Generative Pretrained Transformer (GPT) models showing increasingly effective capabilities. The evolution from GPT-3 \citep{brown2020language} to GPT-4 \citep{openai2023gpt4}, along with the emergence of other LLMs such as PaLM \citep{chowdhery2022palm} and LLaMA \citep{touvron2023llama}, has marked a significant leap in natural language understanding and problem-solving abilities. The generative nature of these models has led to their widespread adoption in numerous application fields. 
Despite their advanced capabilities, LLMs are capable of generating incorrect results \citep{ji2023survey}, an issue that is particularly problematic in applications where precision and dependability are critical, like the biomedical and healthcare fields \citep{azamfirei2023large, nori2023capabilities}.

Existing approaches for improving the correctness of LLMs include prompt engineering \cite{white2023prompt}, retrieval methods \cite{chen2023benchmarking}, domain-specific tuning \cite{wu2023bloomberggpt, nguyen2023brief} among many others \citep{wang2023survey}. While existing methods show varying degrees of improvement on different tasks, in general there lacks a systematic way to efficiently quantify the likelihood of errors in LLM outputs. One strategy involves querying the LLM in various ways to estimate the answer's correctness (e.g., \citealp{manakul2023selfcheckgpt}). However, these approaches are computationally expensive and biased by the LLM itself. 




In many LLM applications, including disease prediction \cite{han2023evaluation}, medical diagnosis \cite{shea2023use}, and question answering (QA) \cite{moore2023assessing}, a concrete and precise answer is desired. For these mission critical tasks, a quantitative error estimation or confidence level for the response is equally important as giving a correct answer itself. However, the text-in text-out nature of the generative language models makes it challenging to estimate the error probability of the answer quantitatively. Although some LLMs provide internal probability scores for the generated tokens, they are poorly calibrated to the true error rate, particularly after applying reinforcement learning with human feedback (RLHF) \citep{ouyang2022training, openai2023gpt4}. 

Our goal in this paper is to address this issue by establishing a systematic way to quantitatively estimate the likelihood of error in LLM answer. We approach this through training an estimator model $h$ via multi-objective optimization, leveraging extensive research in Pareto optimization \citep{pareto1964cours}.
Given the optimized model $h$ and any LLM response, we can then directly estimate the LLM response error rate which we refer to as the Pareto optimal learning assessed risk (POLAR) score (Section \ref{sect-pol}). This framework leverages structured information retrieval from other information sources such as knowledge bases.
We introduce a novel approach that trains a Pareto-optimal probabilistic model $h$ to simultaneously optimize on LLM and align with the external information sources.


Our key contributions are as follows: 

i)  We propose a novel framework using Pareto optimization aligning to the LLM and  multiple external information sources.

ii)  The POLAR score from our framework is shown experimentally to be effective in estimating LLM error rate.

iii)  We demonstrate that POLAR scores can be leveraged to boost an LLM's performance by easily combining with other popular strategies such as self-verification \cite{weng2023large} and retrieval augmented generation (RAG) \cite{chen2023benchmarking}.

\section{Related Work}
Several heuristics have been proposed to reduce error and estimate the confidence level of an LLM response. \citet{wang2022self} used self-consistency to infer the reliability of the answer. \citet{manakul2023selfcheckgpt} proposed SelfCheckGPT as a black-box method to detect hallucination. The chain-of-thought method \cite{wei2022chain} has also been used to indicate potential errors. These approaches are not able to provide a quantitative error estimation that is calibrated to the true error rate, and susceptible to the issue that the model's self-assessment of confidence is inherently biased. The quality of the results is also highly dependent on the prompting strategy.

Producing confidence scores that are well correlated with a model's error rate has been well established in traditional machine learning. The study of model calibration dates back to the seminal work of Platt scaling \citep{platt1999probabilistic}, where a Logistic calibration model is fitted on top of the original model output. Various techniques have been developed afterwards for model calibration, including isotonic regression \citep{zadrozny2002transforming}, temperature scaling \citep{guo2017calibration}, and Bayesian binning \citep{naeini2015obtaining}. For LLM, a contextual calibration method for LLMs was proposed by \citep{zhao2021calibrate}, which adjusts the class balance by taking an ensemble of LLM queries with content-free input. These methods rely on annotated calibration data and access to model's output probabilities that are not always available.

The problem of aggregating multiple sources of information or supervision sources is studied extensively in programmatic weak supervision \citep{zhang2022survey, fu2020fast, varma2019multi, wang2018deep, lang2021self}. Notable works include distant supervision \citep{hoffmann2011knowledge}, crowd-sourcing \citep{raykar2010learning}, data programming \cite{ratner2016data, ratner2017snorkel} and MeTaL, also known as Snorkel, \citep{ratner2019training}.
Most of these works weigh the supervision sources across all examples and combine the multiple sources into a single label per example.
This approach have shown success but also exhibits significant limitations when applied to identifying LLM errors, primarily due to the weighting dilemma, where if the weight assigned to the LLM is too low, the aggregated result can be noisy, and if the LLM weight is too high, the output is dominated by the LLM, making detecting LLM error difficult.
\citet{ruhling2021end} mitigates the weighting problem with instance-dependent weighting, but the expectation maximization procedure demonstrates significant learning variance.

In this work we present a framework to systematically estimate error of an LLM output by simultaneously aligning to multiple information sources while circumventing the weighting dilemma through Pareto optimization.

\section{Methodology}
\label{sect-method}


Our error estimation framework for LLM responses is a two-step process.  
In the first step, we iterate through a corpus of input instances to collect the corresponding LLM responses, while dynamically retrieving heuristic answers from other information sources. In this process, a probabilistic function $h$ is learned that fits the multiple sources in a Pareto optimal manner. In the second step, the optimized $h$ model is used to estimate the error rate of the LLM response on any new input instance, which we refer to as the POLAR score.

\begin{figure*}[ht]
    \centering
    \includegraphics[width=0.75\textwidth]{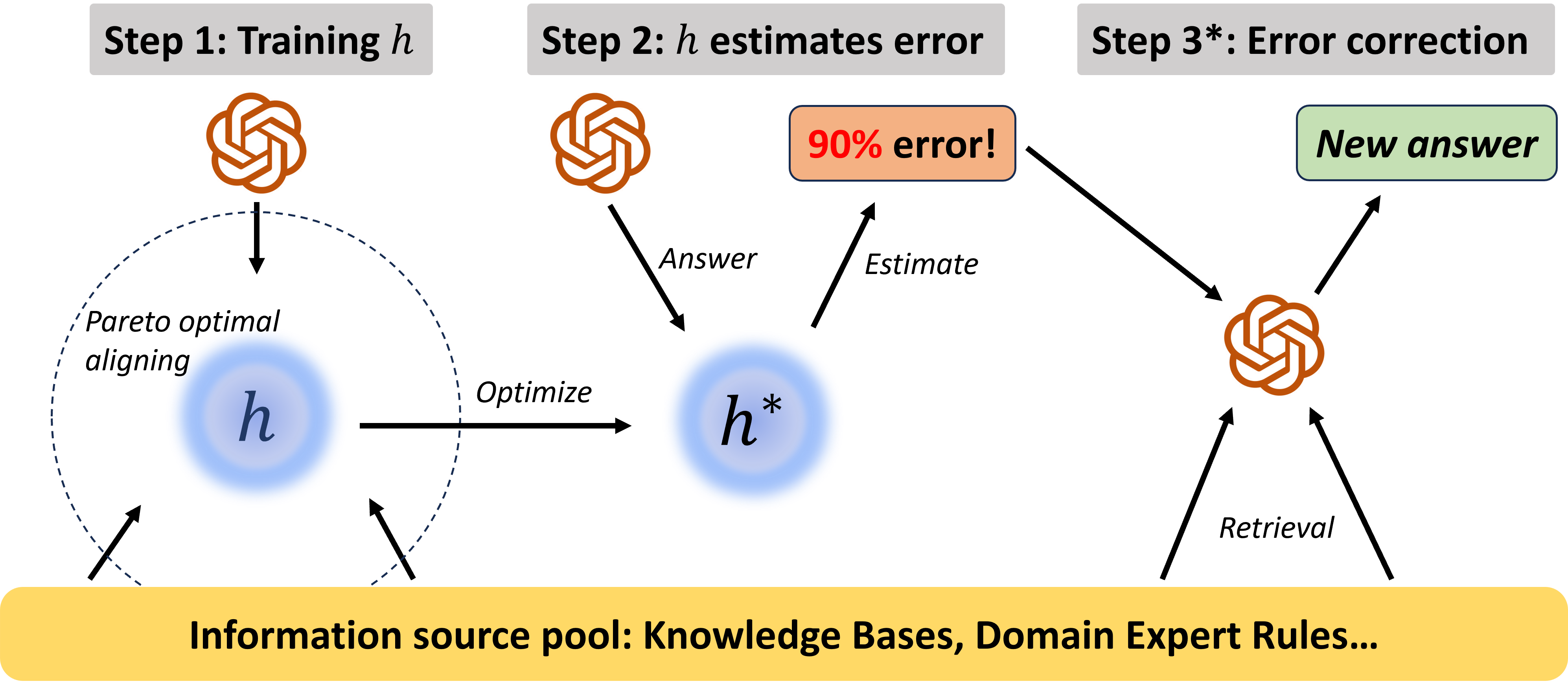}
    \caption{Pareto optimal learning framework for LLM error estimation and correction.}
    \label{fig:framework}
\end{figure*}

After the error estimation step, we also provide an optional third step that strategically re-prompt the LLM based on the POLAR score, and leverage the information retrieved from the information sources in an RAG manner \cite{chen2023benchmarking}. An overview of the framework is shown in Figure \ref{fig:framework}.

\subsection{Problem setup}\label{sect-problem-setup}



Denote the LLM as a function $\mbox{LLM}(x; p)$ where $x \in \mathcal{X}$ is the input text, and $p$ is the user-defined prompt specific to some task. In order to quantify LLM error, we define $\mathcal{Y}$ as the quantized output space (e.g. answer choices in QA, disease type in diagnosis tasks, etc.). Any free-text LLM output is mapped to the quantized space $\mathcal{Y}$ through mapping $\mathcal{Q}$. Now we can define the LLM answer as
\begin{equation}\label{eq-llm}
\Lambda(x) := \mathcal{Q}(\mbox{LLM}(x; p)) \in \mathcal{Y}  \text{,}
\end{equation}
where the output space $\mathcal{Y}$ is of cardinality $K$. Note that the LLM is allowed to state ``unsure'', but in error estimation we only consider the scenario when the LLM explicitly states the answer. Suppose the true answer for input $x$ is $y \in \mathcal{Y}$, estimating the LLM error rate is to estimate the probability $\mathbb{P}(\Lambda(x) \neq y )$. 


To account for other information sources, such as knowledge bases and expert rules, we introduce a pool of $m$ sources. For $j = 1, \cdots, m$, define the triggering function $\mathbb{I}_j(x) \in \{0, 1\}$ indicating if the input $x$ triggers a retrieval from information source $j$. Example triggering includes recognized entities, text patterns, and etc.. Multiple sources can be triggered at the same time.

Once $x$ triggers retrieval from source $j$, i.e. $\mathbb{I}_j(x) = 1$,  the retrieval function $\lambda_j: \mathcal{X} \rightarrow \mathcal{Y}$ represents the answer suggested by information source $j$ for input $x$. Note that the LLM function $\Lambda$ defined in \eqref{eq-llm} is also an information source that maps $\mathcal{X} \rightarrow \mathcal{Y}$. For simplicity, we denote $\lambda_0 := \Lambda$ and $\mathbb{I}_0 = 1$. The answers retrieved from the multiple information sources can conflict with the LLM and among themselves, thus require a proper aggregation, which is the main challenge to solve.

\subsection{Pareto Optimal Learning Assessed Risk}\label{sect-pol}

\paragraph{Step 1: Pareto Optimal Learning}

We propose to learn an optimal function $h: \mathcal{X} \rightarrow K\text{-simplex}$ by simultaneously fitting it to the output of the LLM and other information sources. Incorporating multiple sources helps reduce error dependency on a single source, and corrects the bias from the LLM itself by aligning to external knowledge. The primary challenge here is to design a framework to resolve conflicts among different sources and with the LLM. We do not assume the sources to be independent, as it is often assumed in the weak supervision literature \cite{zhang2022survey}. Instead we adopt the looser assumption that the sources are individually positively correlated with the correct output $y$, better than a random guess. Ideally, a reasonable $h$ should align with each source $j$ if it is triggered for retrieval, which is measured by 

\begin{equation}\label{eq-loss}
\ell_j(h; x) := \mathbb{I}_j(x) \cdot \mathcal{L}(h(x), \lambda_j(x)) \text{,}
\end{equation}
where $\mathcal{L}$ is the cross-entropy loss of $h$ against the information source. Note that $h(x) \in K\text{-simplex}$, which essentially estimates the probability distribution on the output space $\mathcal{Y}$ with cardinality $K$.

In order for $h$ to align with the multiple information sources, mathematically, we want to solve the following multi-objective problem for $j = 0, \cdots, m$:
\begin{equation}\label{eq-fit_end}
\min_{h \in \mathcal{H}} \quad {\mathbb{E}_{x \sim \mathcal{X}}[\ell_j(h; x)]}_{j=0}^m\text{.}
\end{equation}




As the objectives may conflict, we seek an $h^* \in \mathcal{H}$ that is \textit{Pareto optimal}, following multi-objective learning theory \citep{hwang2012multiple} and Pareto optimization \citep{pareto1964cours}.

\begin{definition}[Pareto optimal]\label{def-pareto}
    $h^* \in \mathcal{H}$ is Pareto optimal to $\lambda_0, \cdots, \lambda_m$,  if no $h \in \mathcal{H}$ exists that Pareto dominates $h^*$ in \eqref{eq-fit_end}. $h^*$ must satisfy
\[
\nexists\ h \in \mathcal{H}, \  \text{s.t.} \left\lbrace\begin{array}{l}
    \forall j=0,\cdots, m, \\ \quad \mathbb{E} [\ell_j(h;x)] \leq \mathbb{E} [\ell_j(h^*;x)], \\
   \exists j=0,\cdots, m, \\ \quad  \mathbb{E} [\ell_j(h;x)] < \mathbb{E} [\ell_j(h^*;x)].
  \end{array}\right.
\]
\end{definition}

The Pareto optimization framework effectively manages dependencies between information sources. For example, a Pareto optimal $h$ remains unaffected by the arbitrary duplication of a source. However, finding Pareto optimal solutions remains challenging in the multi-objective optimization literature. In this paper, we adopt one of the standard approaches to scalarize the multiple objectives \cite{hwang2012multiple}, and solve
\begin{equation}\label{eq-pareto}
\min_{h\in \mathcal{H}} \mathbb{E}_{x \sim \mathcal{X}}[G(\vec{\ell})],
\end{equation}
where $\vec{\ell} = (\ell_0, \cdots, \ell_m)$ and $G$ is a Pareto aggregator as defined below.

\begin{definition}[Pareto aggregator]\label{def-scale}
    $G: \mathbb{R}_+^{m+1} \rightarrow \mathbb{R}_+$ is a \textit{Pareto aggregator} if it satisfies:
\begin{itemize}
    \item $G$ is convex, and
    \item $G(\vec{\ell}_j') < G(\vec{\ell})$ if $\ell_j' < \ell_j\ \forall\ j = 0, \cdots, m$, where $\vec{\ell}_j'=(\ell_0, \cdots, \ell_j', \cdots, \ell_m)$.
    
\end{itemize}
\end{definition}

In this study, we explore four different types of aggregators:
\begin{itemize}
    \item Linear: $G(\vec{\ell}) = \sum_{j=0}^m \ell_j = \|\vec{\ell}\|_1$,
    \item Quadratic: $G(\vec{\ell}) = \left(\sum_{j=0}^m \ell_j \right)^2 = \|\vec{\ell}\|_1^2$,
    \item Euclidean norm: $G(\vec{\ell}) = \sqrt{\sum_{j=0}^m \ell_j^2} = \|\vec{\ell}\|_2$,
    \item Chebyshev: $G(\vec{\ell}) = \max_{j=0}^m \ell_j = \|\vec{\ell}\|_{\infty}$.
\end{itemize}

The nonlinear aggregator shapes the optimization in Equation~\eqref{eq-pareto} differently through Jensen's inequality.
While the first three aggregators qualify as Pareto aggregators, the Chebyshev aggregator does not meet the definition criteria, serving as a comparative element in our experiment.


With Definitions \ref{def-pareto} and \ref{def-scale}, we propose finding an optimal solution $h^*$ by solving Equation \eqref{eq-pareto}, which can be done via standard stochastic gradient descent algorithms such as Adam \citep{kingma2014adam}. The solution is guaranteed by the following Theorem, with a detailed proof in Appendix \ref{sect-proof}.
\begin{theorem}\label{theorem-pareto}
    Suppose $G$ is a Pareto aggregator as in Definition \ref{def-scale}, solving the problem in Equation \ref{eq-pareto} approximates a Pareto optimum by minimizing the upperbound.
\end{theorem}

\paragraph{Step 2: POLAR Score Estimation}

Once a optimal solution $h^*$ is found, we can estimate the error rate of $\Lambda(x)$ for any new input $x \in \mathcal{X}$, by selecting the probability in $h^*(x)$ that corresponds to $\Lambda(x)$, denoted $h^*(x)[\Lambda(x)]$, and compute a risk score
\begin{align}
    \zeta(x, \Lambda(x); h^*)&=\mathbb{P}_{Y\sim {h^*(x)}}(\Lambda(x) \neq Y)  \nonumber \\
    &= 1-h^*(x)[\Lambda(x)]
    \text{,} \label{eq-proba}
\end{align}
where $\mathbb{P}_{Y\sim {h^*(x)}}$ represents the probability distribution of $Y$ as estimated by $h^*(x)$. We refer to $\zeta(x, \Lambda(x); h^*)$ as the Pareto optimal learning assessed risk (POLAR) score. 

Algorithm \ref{alg:polar_algorithm} summarizes the entire process. 







\begin{algorithm}[ht]
\caption{POLAR score estimation}\label{alg:polar_algorithm}
Step 1: Training estimator $h^*$
\begin{algorithmic}[1]
\State \textbf{Input:} LLM with prompt $p$, answer mapping $\mathcal{Q}$, retrieval triggering functions $\mathbb{I}_{1:m}$, information sources $\lambda_{1:m}$, training inputs $x_{1:n}$, Pareto aggregator $G$, and initialized $h \in \mathcal{H}$.
\For{$i = 1$ to $n$}
\State $\Lambda(x_i)=\mathcal{Q}(\text{LLM}(x_i;p))$
\State $\text{L}_{\text{LLM}} = \mathcal{L}(h(x_i), \ \Lambda(x_i))$
\For{$j = 1$ to $m$}
\State $\text{L}_j = \mathbb{I}_j(x_i) * \mathcal{L}(h(x_i), \ \lambda_j(x_i))$
\EndFor
\State Update $h$ with SGD iteration of $\min_h G(\text{L}_{\text{LLM}}, \ \text{L}_1, \cdots, \text{L}_m)$.
\EndFor

\State \textbf{Output:} Optimized $h^*$. 
\end{algorithmic}

Step 2: Estimation with $h^*$.
\begin{algorithmic}[1]
\State \textbf{Input:} New input $x$, optimized $h^*$, LLM with prompt $p$, answer mapping $\mathcal{Q}$.
\State LLM answer $\Lambda(x)=\mathcal{Q}(\text{LLM}(x;p))$
\State \textbf{Output:} POLAR score $\zeta(x, \Lambda(x); h^*) = 1-h^*(x)[\Lambda(x)]$ in Equation \eqref{eq-proba}.
\end{algorithmic}
\end{algorithm}


\subsection{Step 3*: Error Correction with POLAR}\label{sect-dynamic}
Identifying LLM responses with a higher risk of error presents an opportunity to efficiently correct the errors and improve the final accuracy. In most of the applications, the POLAR score itself is sufficient to facilitate human-in-the-loop intervention. Here we provide an optional Step 3, which easily connects the POLAR score to other prompting strategies to correct the error automatically. In this setting, the information sources also serve as additional input to the LLM to enhance its answer. In this paper, we propose two dynamic prompting strategies to illustrate correcting LLM errors using the POLAR score $\zeta(x, \Lambda(x); h^*)$.

\paragraph{Dynamic self-verification}
Pick risk threshold $\delta$. For any input $x$ and LLM answer $\Lambda(x)$, if $\zeta(x, \Lambda(x); h^*) > \delta$, simply ask the LLM to self-verify its previous answer.

\paragraph{POLAR-assisted RAG}
Pick risk threshold $\delta$. For any input $x$ and LLM answer $\Lambda(x)$, if $\zeta(x, \Lambda(x); h^*) > \delta$, retrieve information from all sources that are triggered, i.e. $\mathbb{I}_j(x) = 1$. Provide the answer suggested by the information sources $\lambda_j(x)$ and the description of the sources to the LLM, and generate the revised answer. Algorithm 2 outlines the POLAR-assisted RAG. We provide detailed prompting design description in the Appendix \ref{sect-prompt}. 

\begin{algorithm}[ht]
\caption{POLAR-assisted RAG}\label{alg:dynamic_ss}
\begin{algorithmic}[1]
\State \textbf{Input:} Input $x$, optimized $h^*$, LLM with prompt $p$, answer mapping $\mathcal{Q}$, retrieval triggering functions $\mathbb{I}_{1:m}$, information sources $\lambda_{1:m}$, text description of the sources $d_{1:m}$, and user-defined POLAR threshold $\delta$.
\State Initial LLM answer $\Lambda(x)=\mathcal{Q}(\text{LLM}(x;p))$
\State Compute POLAR score \[\zeta(x, \Lambda(x); h^*) = 1-h^*(x)[\Lambda(x)]\]
\If{$\zeta(x, \Lambda(x); h^*) > \delta$}
\State Initialize a followup prompt $p'$.
\For{$j = 1$ to $m$}
\If{$\mathbb{I}_j(x) = 1$}
\State Get source answer $\lambda_j(x)$.
\State Get source description $d_j$
\State $p' = p' +  \lambda_j(x) + d_j$.
\EndIf
\EndFor
\State $\Lambda'(x)=\mathcal{Q}(\text{LLM}(x; p'))$
\EndIf
\State \textbf{Output:} Updated answer $\Lambda'(x)$
\end{algorithmic}
\end{algorithm}
\section{Experiments}
\label{sect-exp}

\paragraph{Dataset}

In order for the results to be reproducible, it is essential that both the datasets themselves and the associated extra information sources are fixed in the benchmarks. We leverage the publicly available datasets collected by \cite{zhang2021wrench} and evaluate on four different NLP tasks: CDR \citep{li2016biocreative}, ChemProt \citep{krallinger2017overview}, SemEval \citep{hendrickx2019semeval}, and SMS \citep{almeida2011contributions}. The pre-defined supervision functions in these benchmarks serve as the information sources $\lambda_j$ in addition to the LLM output $\Lambda$. We do not use any of the ground truth in the training sets, and only use the ground truth labels on test set for evaluation of the LLM error rate and error correction performance. Our dataset selection covers the following aspects:
\begin{itemize}
\item Domain: General domain tasks(SemEval, SMS) that requires general instruction following and reasoning, as well as biomedical domain tasks (CDR, ChemProt) that requires domain knowledge.

\item Difficulty: Tasks that are easy for advanced LLMs, like SMS (99\% F1 for GPT-4), and tasks that are still challenging, like CDR (74\% F1 for GPT-4), ChemProt (42\% F1 for GPT-4), and SemEval (67\% F1 for GPT-4). We illustrate that our framework can effectively estimate the error rate even when the LLM is already nearly perfect on the task.
\end{itemize}

\paragraph{Prompt design}
To maximize LLM capabilities, we carefully design prompts for each problem, clarifying the problem setting, knowledge background, input/output structure, and instructions for stating "unsure". Detailed prompt information is provided in Appendix \ref{sect-prompt}.

\paragraph{Information sources}
The information sources in these datasets were created by human experts that utilize knowledge bases (e.g. the Comparative Toxicogenomics Database; \citealp{davis2021comparative}), textual patterns, and a combination of the two \citep{ratner2017snorkel, yu2021fine, zhou2020nero, awasthi2020learning}. We refer to the benchmark by \cite{zhang2021wrench} for detailed description.

\paragraph{Optimization}
The Pareto optimal $h(x)$ is obtained by solving Equation \ref{eq-pareto}, and we choose the quadratic aggregator for $G$ and the BERT model \citep{DBLP:journals/corr/abs-1810-04805} for $h$. For the biomedical CDR and ChemProt, we choose BiomedBERT \citep{pubmedbert} for $h$. Ablations of model architectures and aggregators are examined in Section \ref{sect-ablation}. Details of the training setup is described in Appendix \ref{sect-train}.


\begin{figure*}[ht]
    \centering
    \begin{subfigure}[b]{0.3\textwidth}
        \includegraphics[width=\textwidth, height=\textheight, keepaspectratio, trim={0 0 0 0.35in},clip]{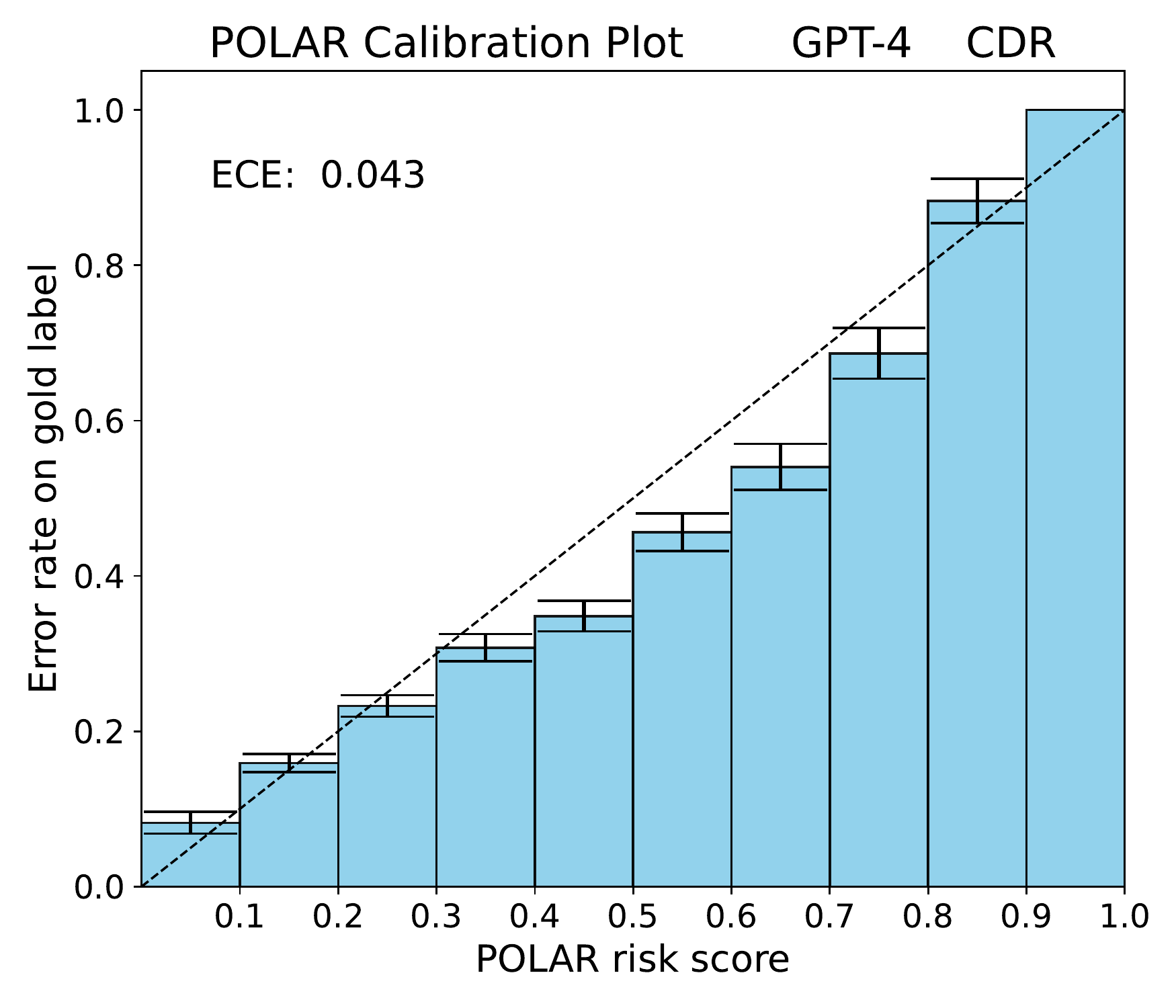}
        \caption{Error calibration curve}
        \label{fig:cal_curve}
    \end{subfigure}
    \hfill
    \begin{subfigure}[b]{0.3\textwidth}
        \includegraphics[width=\textwidth, height=\textheight, keepaspectratio, trim={0 0 0 0.37in},clip]{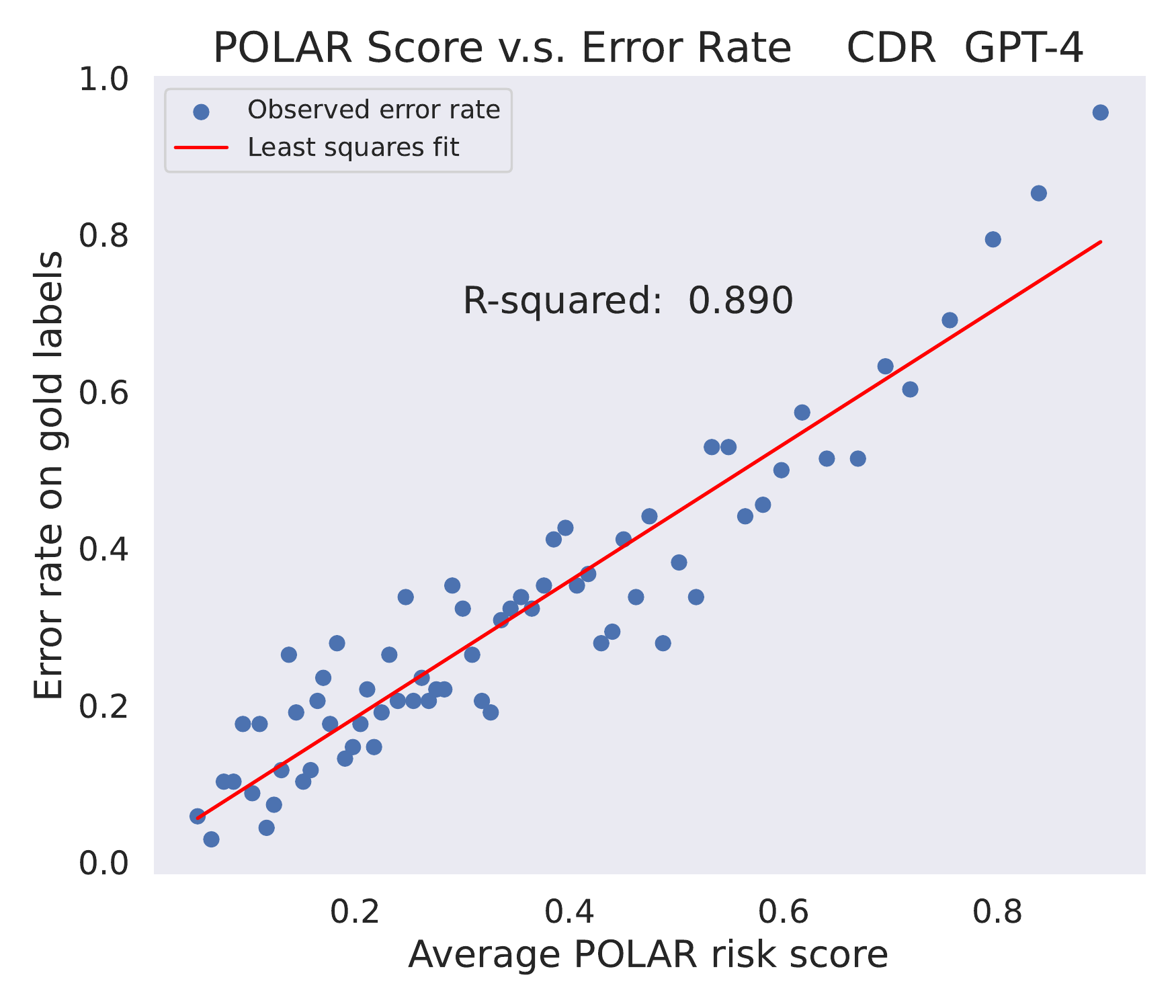}
        \caption{Correlation with error rate}
        \label{fig:corr}
    \end{subfigure}
    \hfill
    \begin{subfigure}[b]{0.37\textwidth}
        \includegraphics[width=\textwidth, height=\textheight, keepaspectratio, trim={0 0 0 0.42in},clip]{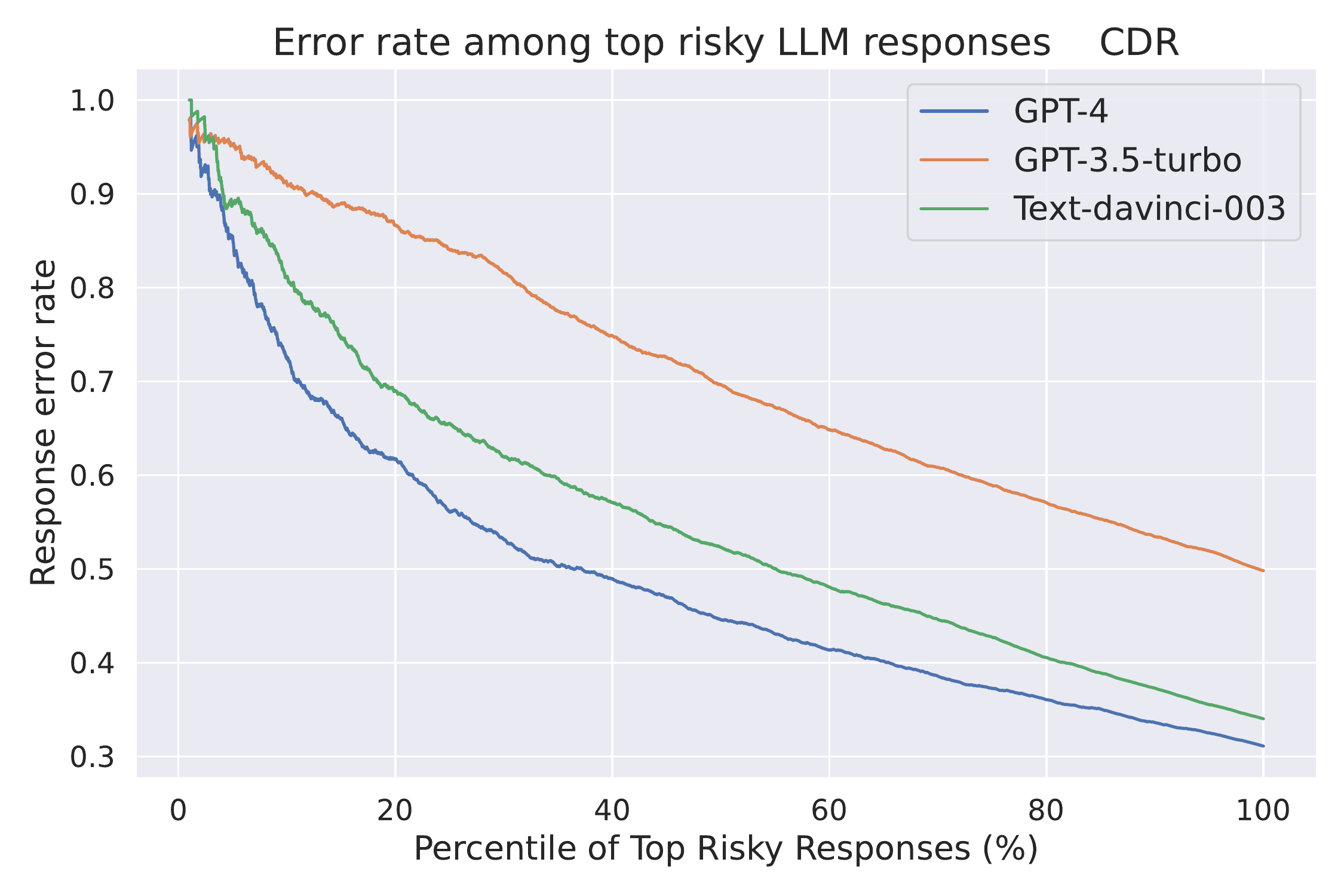}
        \caption{LLM error detection}
        \label{fig:top_risk}
    \end{subfigure}
    \caption{LLM error estimation using the POLAR score. (a) The LLM response error rate vs ten equal-interval POLAR score bins. (b) The POLAR scores are sorted and then binned where each bin contains 100 examples. The average of the LLM errors and POLAR scores are plotted for each bin. The last bin with the top POLAR scores may have less than 100 examples. (c) shows the average LLM error rate vs top percentile POLAR score examples.}
    \label{fig-calibration}
\end{figure*}

\subsection{LLM Error Estimation}

We present the LLM error estimation results using the POLAR score and compare them with existing methods as baselines. 

\subsubsection{Evaluation metrics} As the primary interest is in how well the LLM error rate is estimated, we report evaluation results measured in expected calibration error (ECE) \cite{naeini2015obtaining} that is widely used in assessing model confidence calibration as well as the coefficient of correlation $R^2$. An ECE of $0$ indicates perfectly estimated error rates, or equivalently a perfectly calibrated model $h$ in our framework.


For each dataset, the training split is used as the input examples $x_1,\cdots,x_n$ along with the predefined information sources $\lambda_i$ to learn $h^*$. The test split and its ground truth labels are used obtain to $\Lambda(x)$, compute the true LLM error rate, and evaluate $h^*$ measured by ECE and $R^2$.

Figure \ref{fig-calibration} displays POLAR score calibration for GPT-4 on the CDR chemical-disease relation extraction task. The calibration curve (Figure \ref{fig:cal_curve}) shows that the POLAR score reliably estimates the true probability of LLM error rates. Figure \ref{fig:corr} demonstrates a high correlation between the POLAR score and the true error rate. Figure \ref{fig:top_risk} reveals that responses with the highest POLAR scores are most prone to errors, with the top scores indicating nearly a 100\% error rate.

\subsubsection{Baseline methods}
We compare the proposed POLAR score with the following baseline error estimation approaches. We divide them based on if they utilized the information sources that were used in our framework.

\paragraph{Without information sources}
We estimate the error rate through LLM ensemble: 
\begin{itemize}
    \item[1.] Query the LLM multiple times with the same input $x$ to sample responses. 
    \item[2.] Estimate the probability where the answer is different from the most frequent one, and use it as the estimation for error rate on input $x$.
\end{itemize} 
As this approach is extremely expensive, we only evaluated this for GPT-3.5-turbo on the CDR dataset. The estimated ECE in this experiment is 0.4466, which is far worse than other approaches in Table \ref{calibration-table}. Therefore, we focus on comparison with other baseline methods in the rest of this section.

\paragraph{With information sources} The following methods utilize the exact same information sources as in our framework. They differ in how the answers from the multiple sources and the LLM were combined together.
\begin{itemize}
    \item Snorkel \citep{ratner2019training}: a weak supervision method combining multiple supervision sources via matrix completion. Snorkel model fitted on training set is use to give class probabilities for LLM error rate. We use the class probability given by the fitted model to estimate LLM error rate.
    \item WeaSEL \citep{ruhling2021end}: A state-of-the-art weak supervision framework that fits a neural network using the LLM and weak labels in an end-to-end manner. We use the class probability given by the fitted model to estimate LLM error rate.
    \item Majority vote: A common method that estimates class probability according to the voted ratios among all information sources.
    
\end{itemize}


\begin{table*}[ht]
  \caption{LLM error rate estimation performance, using the POLAR score and other methods, measured in ECE and $R^2$. The best entries (low ECE, high $R^2$) from each row are highlighted in bold.}
  \label{calibration-table}
  \centering
  \begin{tabular}{llllllllll}
    \toprule
    Task & LLM & \multicolumn{2}{c}{\textbf{POLAR}} & \multicolumn{2}{c}{Snorkel} & \multicolumn{2}{c}{WeaSEL} & \multicolumn{2}{c}{Majority vote} \\
    \midrule
    &  &  ECE & $R^2$ & ECE & $R^2$ & ECE & $R^2$ & ECE & $R^2$ \\
    \cmidrule(r){3-4} \cmidrule(r){5-6} \cmidrule(r){7-8} \cmidrule(r){9-10}
    \textbf{CDR} & GPT-4            & \textbf{0.043} & \textbf{0.890} &  0.167 & 0.299 &   0.146 & 0.387 &   0.145 & 0.348 \\
    & GPT-3.5-turbo     & \textbf{0.046} & 0.934 &  0.164 & 0.320 &   0.081 & \textbf{0.942} &     0.182 & 0.540 \\
    & Text-davinci-3 & \textbf{0.055} & \textbf{0.907} &  0.154 & 0.371 &   0.135 & 0.877 &     0.149 & 0.450 \\
    \midrule
    \textbf{ChemProt} & GPT-4           & \textbf{0.035} & \textbf{0.934} &  0.182 & 0.510 &   0.278 & 0.885 &     0.233 & 0.244 \\
    & GPT-3.5-turbo     & \textbf{0.048} & \textbf{0.944} &  0.228 & 0.625 &   0.219 & 0.922  &   0.282 & 0.031 \\
    & Text-davinci-3 & \textbf{0.051} & \textbf{0.917} &  0.218 & 0.700 &    0.213 & 0.846 &    0.279 & 0.307 \\
    \midrule
    \textbf{SemEval} & GPT-4            & 0.079 & \textbf{0.916} &  \textbf{0.068} & 0.714 &    0.612 & 0.626 &    0.115 & 0.379 \\
    & GPT-3.5-turbo     & \textbf{0.047} & \textbf{0.963} &  0.150 & 0.821 &   0.345 & 0.890 &     0.277 & 0.208 \\
    & Text-davinci-3 & \textbf{0.067} & \textbf{0.950} &  0.119 & 0.796 &   0.455 & 0.784 &     0.242 & 0.396 \\
    \midrule
    \textbf{SMS} & GPT-4           & \textbf{0.014} & \textbf{0.980} &  0.244 & 0.089 &  0.409 & 0.345 &      0.588 & 0.091 \\
    & GPT-3.5-turbo     & \textbf{0.041} & \textbf{0.963} &  0.075 & 0.202 &   0.286 & 0.731 &     0.148 & 0.006 \\
    & Text-davinci-3 & \textbf{0.023} & \textbf{0.943} &  0.201 & 0.053 &   0.420 & 0.238  &   0.325 & 0.091 \\
    \bottomrule
  \end{tabular}
\end{table*}

Table \ref{calibration-table} compares POLAR score performance in LLM error calibration against baseline methods. We report the results spanning the four datasets and three LLMs (GPT-4, GPT-3.5-turbo, and text-davinci-003). The proposed POLAR score consistently outperforms other methods. Among the baseline methods, Snorkel, WeaSEL, and LLM distilled model can achieve top or close-to-top performance in some cases under specific metric, but lack the consistency to deliver stable calibration for different LLMs on different tasks. In comparison, the proposed POLAR score is consistently well-calibrated to the true error rate.

\begin{figure}[h!]
    \centering
    \begin{subfigure}[b]{0.45\textwidth}
        \includegraphics[width=\textwidth]{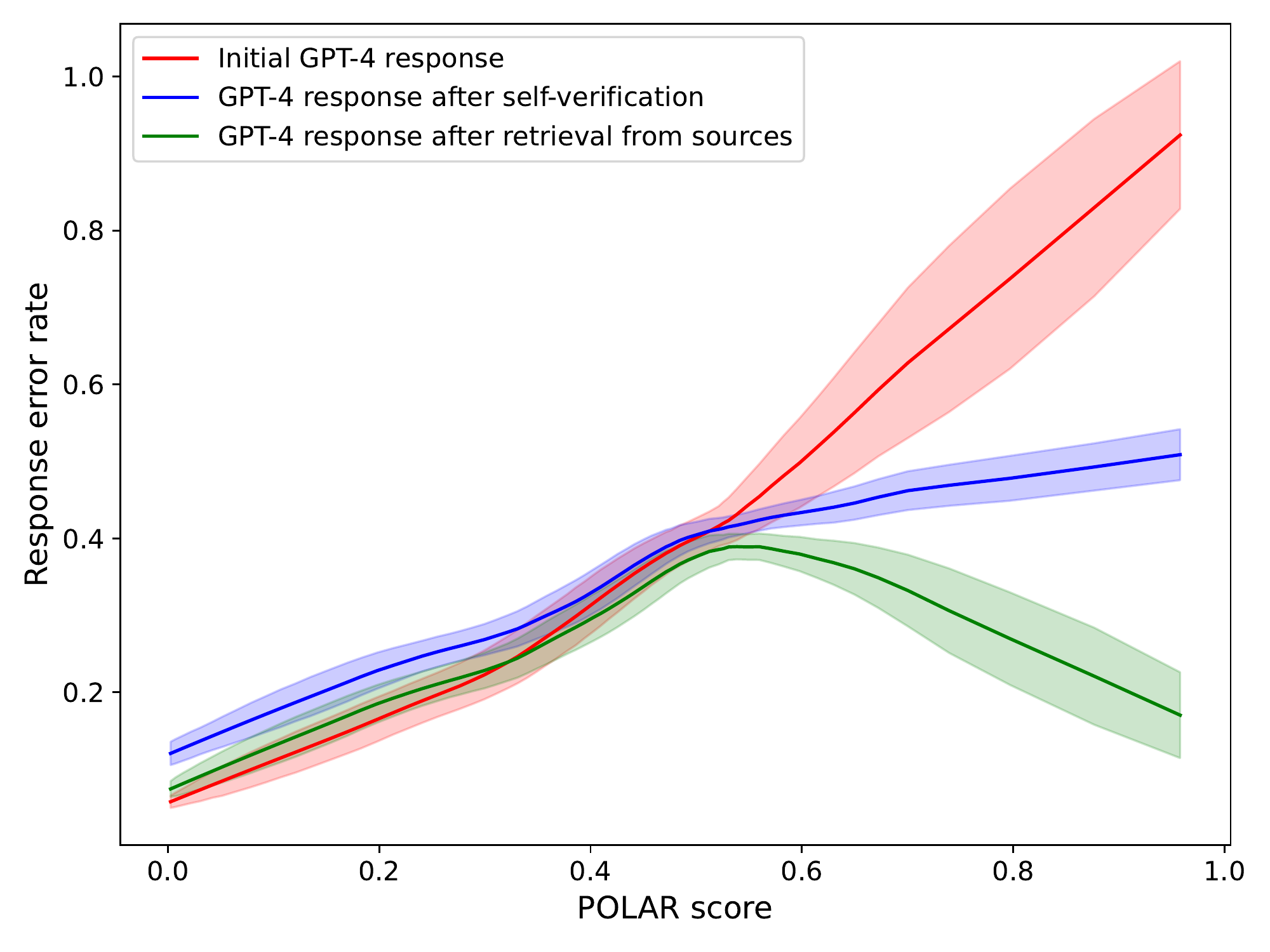}
        \caption{Error rate reduction conditioning on POLAR score.}
        \label{fig:dynamic_polar}
    \end{subfigure}
    \hfill
    \begin{subfigure}[b]{0.45\textwidth}
        \includegraphics[width=\textwidth]{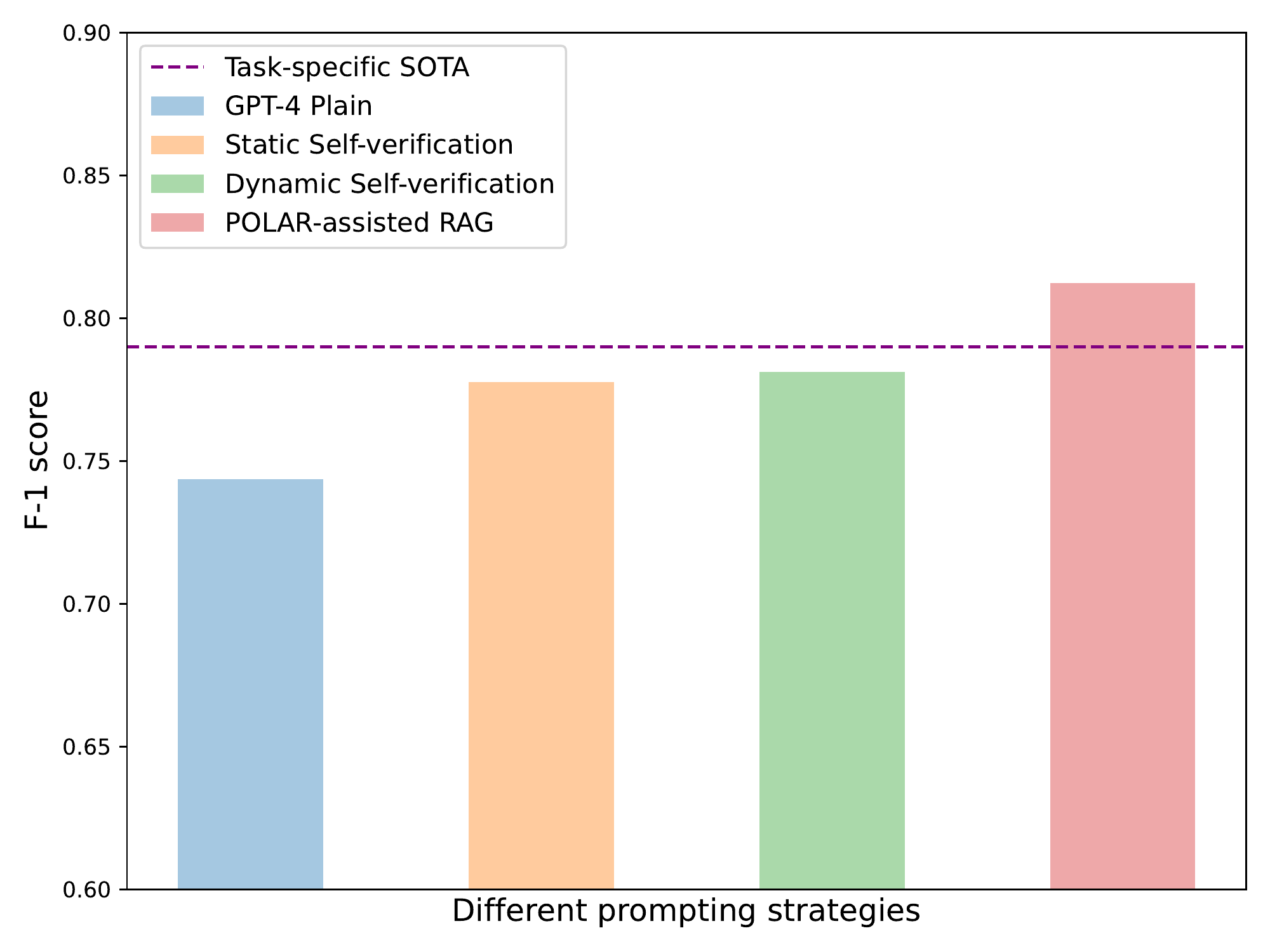}
        \caption{Dynamic prompting performance}
        \label{fig:dynamic_improve}
    \end{subfigure}
    \caption{(a) shows the GPT-4 error rate before and after re-prompting, as plotted against the POLAR score. (b) shows the performance improvement using the two dynamic prompting strategies in Section \ref{sect-dynamic}.}
    \label{fig-followup}
\end{figure}

\subsection{Improved LLM performance with POLAR-assisted dynamic prompting}
We investigate the utility of the POLAR score through dynamic prompting (Section \ref{sect-dynamic}) to rectify LLM errors. In this experiment, we focus only on the CDR dataset and GPT-4 and GPT-3.5-turbo models.

\begin{table*}[h!]
  \caption{LLM error estimation performance with and without external information sources.}
  \label{extra_info-table}
  \centering
  \begin{tabular}{lllllllll}
    \toprule
Dataset & \multicolumn{2}{c}{CDR} & \multicolumn{2}{c}{ChemProt} & \multicolumn{2}{c}{SemEval} & \multicolumn{2}{c}{SMS}   \\
\midrule

&    ECE &     $R^2$ &    ECE &     $R^2$ &    ECE &     $R^2$ &    ECE &     $R^2$ \\
\cmidrule(r){2-3} \cmidrule(r){4-5} \cmidrule(r){6-7} \cmidrule(r){8-9} 
POLAR (with source) & \textbf{0.043} & \textbf{0.890} & \textbf{0.035} & \textbf{0.934} & 0.079 & 0.916 & 0.014 & \textbf{0.980} \\
Without Sources  & 0.164 & 0.592 & 0.216 & 0.766 & \textbf{0.063} & \textbf{0.947} & \textbf{0.013} & 0.977 \\
\bottomrule
\end{tabular}
\end{table*}

\begin{table*}[h!]
  \caption{Average LLM error estimation performance for different loss aggregator and $h$ modeling choices.}
  \label{loss_model-table}
  \centering
  \begin{tabular}{lllllllll}
    \toprule
Aggregator & \multicolumn{2}{c}{Linear} & \multicolumn{2}{c}{Quadratic} & \multicolumn{2}{c}{Euclidean norm} & \multicolumn{2}{c}{Chebyshev}   \\
\midrule

Architecture &    ECE &     $R^2$ &    ECE &     $R^2$ &    ECE &     $R^2$ &    ECE &     $R^2$ \\
\cmidrule(r){2-3} \cmidrule(r){4-5} \cmidrule(r){6-7} \cmidrule(r){8-9} 

BERT & 0.0625 & 0.9273 & \textbf{0.0458} & \textbf{0.9366} & 0.0549 & 0.9003 & 0.0711 & 0.8260 \\
MLP  & \textbf{0.0555} & \textbf{0.9392} & 0.0974 & 0.9188 & 0.0691 & 0.9302 & 0.0775 & 0.8934 \\
LR   & \textbf{0.0641} & \textbf{0.9360} & 0.1072 & 0.9020 & 0.0766 & 0.9288 & 0.0948 & 0.8813 \\
\bottomrule
\end{tabular}
\end{table*}

To understand the advantage of dynamic prompting, we first examine the effect of LLM error rate using \textit{static} prompting. That is, ignoring the POLAR score, we persistently follow up the initial LLM response with another prompt, either using the self-verification or providing suggested answers from the information sources. The results are shown in Figure \ref{fig:dynamic_polar} where the LLM error rate is plotted with respect to the POLAR score. We observe that the LLM error rate decreases with static follow-up prompts when the POLAR score is high, around $>0.5$. However, the LLM error rate increases slightly with follow-up prompting when the POLAR score is low, around $<0.5$. This suggests the utility of dynamic prompting -- follow up the initial LLM response with a prompt only when the LLM is more likely to make a mistake, i.e. when the POLAR score is high.

We apply dynamic prompting to the same problem and set $\delta=0.5$. The results are shown in Figure \ref{fig:dynamic_improve}. We see that the POLAR-assisted dynamic prompting increases the GPT-4 performance. The POLAR-assisted RAG strategy has a larger increase due to incorporating additional information sources $\lambda_j$ in the follow-up prompt. We note that GPT-4 with POLAR-assisted RAG outperforms state-of-the-art supervised task-specific model \cite{xiao2021sais}. 

\section{Ablations}
\label{sect-ablation}

\paragraph{External sources of information}
 To illustrate the importance of these additional sources, we examine the performance without them by removing all $\lambda_j$ for $j=1, \cdots, m$, and only take $\lambda_0 := \Lambda$ as the LLM itself in the framework. Note that in that sense, solving Equation \eqref{eq-fit_end} becomes a single-objective learning problem, which essentially is equivalent to distilling a small model $h$ from $\Lambda$. We show the error estimation comparison for GPT-4 in Table \ref{extra_info-table}. We can see that the absence of additional information sources generally leads to inconsistent error estimation capabilities, as they are essential to prevent $h$ from overfitting to LLM responses. This is especially obvious for the biomedical domain tasks CDR and ChemProt, where domain knowledge is critical in error estimation.

\paragraph{Pareto loss scalarizer and harmonizer}
The effectiveness of different loss aggregators and $h$ modeling choices is encapsulated in Table \ref{loss_model-table}, where we present the ECE and $R^2$ measures averaged across three LLMs (GPT-4, GPT-3.5-turbo, and text-davinci-003) for the four tasks. We observe that the nonlinear quadratic loss aggregator, when combined with BERT finetuning, delivers superior error estimation performance. Conversely, simpler models, such as Multi-Layer Perceptron (MLP) and Logistic Regression (LR), achieve optimal results with a basic linear aggregator. Notably, the Chebyshev aggregator consistently underperforms across various scenarios. This empirical evidence lends support to Theorem \ref{theorem-pareto}, validating the necessity of a Pareto loss aggregator.

\section{Conclusion}
We presented a novel framework for LLM error estimation using Pareto optimal learning. The error estimator learned in our framework aligns with the LLM and other information sources Pareto optimally. We showed experimentally that the proposed POLAR score is well calibrated with the LLM error rate evaluated on ground truth, ensuring reliable error estimation. We proposed two POLAR-assisted dynamic prompting strategies, and showed that POLAR-assisted RAG enhances GPT-4's performance, surpassing state-of-the-art task-specific model. This development marks a substantial advancement in the application of LLM, providing an effective method to both estimate and reduce LLM errors.


%
%

\newpage

\section*{Limitations}


According to Theorem \ref{theorem-pareto}, we approximate a Pareto optimum by minimizing its upperbound. However, due to the complexity of modeling the errors present in an LLM response and other information sources, we cannot offer further theoretical guarantees of how close an approximated solution is to a Pareto optimum. Nonetheless, we empirically demonstrate in Section \ref{sect-exp} that the approximated solution $h^*$ closely estimates the LLM error rate.

In this work, we require external information sources for retrieval. While they are essential for injecting knowledge in addition to the LLM, we admit that not they are not available for every applications. Our framework also focuses on clearly defined error cases where there is a concrete answer. We envision future work to expand into areas where the LLM output is more ambiguous.



\section*{Ethics Statement}
The motivation of this work is to reduce errors produced by LLMs so that LLMs can be beneficial to users and applications by providing factual and accurate responses. Potential risk includes improper use of the error estimator.


\bibliography{main}
\bibliographystyle{acl_natbib}

\appendix
\counterwithin{theorem}{section}
\counterwithin{proposition}{section}
\counterwithin{definition}{section}

\section{Proof of Theorems}
\label{sect-proof}

\subsection{Proof of Theorem \ref{theorem-pareto}}
\begin{proof}
    For convenience, let's denote 
    \[u_j(h) := \mathbb{E} [\ell_j(h(x), \lambda_j(x))], \quad j=0,1, \cdots, m.
    \]
    We first show that any $h^*$ minimizing $G(u_0, u_1, \cdots, u_m)$ is Pareto optimal.

    Proof by contradiction. Suppose $h^*$ is not Pareto optimal. Then there must exist some $h' \in \mathcal{H}$ Pareto dominating $h^*$. Without loss of generality, let's assume $u_j(h') < u_j(h^*)$, and $u_k(h') \leq u_k(h^*)$, $\forall k \neq j$. Then according to Definition \ref{def-scale} of Pareto aggregator, 
    \begin{align}
        & G(u_0(h'), \cdots, u_j(h'), \cdots, u_m(h'))  \\
        \leq & G(u_0(h^*), \cdots, u_j(h'), \cdots, u_m(h^*))  \\
        < & G(u_0(h^*), \cdots, u_j(h^*), \cdots, u_m(h^*)),
    \end{align}
    which contradicts the assumption that $h^*$ is the minimizer for 
    \[
    G(u_0(h), \cdots, u_j(h), \cdots, u_m(h)).
    \]

    Therefore, the original statement is true, and minimizing the objective 
    \begin{equation}
        \label{eq-obj-ap}
        \min_{h\in \mathcal{H}} \quad G(\mathbb{E}\ell_0(h), \cdots, \mathbb{E}\ell_j(h), \cdots, \mathbb{E}\ell_m(h))
    \end{equation}
    gives a Pareto optimum.

Next, we use Jensen's inequality to upperbound this objective with the objective in problem \ref{eq-pareto}. Using the fact that $G$ is convex, we apply Jensen's inequality and get
\begin{align}
    & G(\mathbb{E}\ell_0(h), \cdots, \mathbb{E}\ell_j(h), \cdots, \mathbb{E}\ell_m(h)) \\
    \leq \quad & \mathbb{E}[G(\ell_0(h), \cdots, \ell_j(h), \cdots, \ell_m(h))].
\end{align}
    Therefore, solving the problem in Equation \ref{eq-pareto} approximates Pareto optimal harmonizer by upperbounding Equation \ref{eq-obj-ap}.
\end{proof}

\bigskip

\section{Weights for Rebalancing the Sources}
\label{sect-weight}

In our experiments, we explored four different types of scalarization functions, namely:
\begin{itemize}
    \item Linear aggregator: $G(\ell_0, \ell_1, \cdots, \ell_m) := \sum_{j=0}^m w_j\ell_j$.
    \item Quadratic aggregator: $G(\ell_0, \ell_1, \cdots, \ell_m) := \left(\sum_{j=0}^m w_j\ell_j \right)^2$.
    \item Euclidean norm aggregator: $G(\ell_0, \ell_1, \cdots, \ell_m) := \| \left(w_0 \ell_0, w_1 \ell_1, \cdots, w_m \ell_m \right)\| $.
    \item Chebyshev aggregator: $G(\ell_0, \ell_1, \cdots, \ell_m) := \max_{j=0}^m w_j\ell_j$.
\end{itemize}
The weights $w_j \in \mathbb{R}$ are parameters of $G$. In the main text of the paper, we fixed to equal weights $\vec{w} = \vec{1}$. Here we introduce three approaches to determine the weighting if necessary.

\paragraph{Equal Weight}
The simplest weighting scheme of
\[
w_0 = w_1 = \cdots = w_m = \frac{1}{m+1}
\]
gives nice performance in practice, and is the method we used for the results in the main body of the paper. The nonlinear Pareto aggregators have the ability to balance the sources even under equal weights. Practically, We recommend starting with equal weight. \\

In the case that the supervision sources are highly correlated, or when the quality of the sources varies a lot, we propose the following two approaches utilizing the correlation in prediction residual. \\

\paragraph{Maximal Eigenvalue of Residual Correlation}
Suppose we have a pilot harmonizer $h_0 \in \mathcal{H}$, which can usually be obtained from minimizing a Pareto aggregator with equal weight, it gives predicted distribution $\Vec{p}(x) \in \mathbb{R}^{|\mathcal{Y}|}$ for any input $x \in \mathcal{X}$, where
\[
p_c(x) := \mathbb{P}(h_0(x) = c).
\]
For any source $0 \leq j \leq m$, denote the one-hot vector $\Vec{\lambda}_j(x) \in \mathbb{R}^{|\mathcal{Y}|}$ as:
\[
\lambda_{j,c}(x) = \left\lbrace \begin{array}{ll}
  1   &  \mbox{if } \lambda_j(x) = c, \\
  0   &  \mbox{otherwise.}
\end{array}\right.
\]

The prediction residual is defined as
\[
\Vec{r}_j(x) := \Vec{\lambda}_j(x) - \Vec{p}(x),
\]
which accounts for the information source label for $x$ that is unexplained by the harmonizer $h_0(x)$.  \\

In order to rebalance the sources, we consider the correlation matrix $C$ between the prediction residual $\Vec{r}_j$'s. Specifically, let the covariance be
\[
\Sigma_{ij} := \mathbb{E}_{x \sim \mathcal{X}}[\Vec{r}_i(x) \cdot \Vec{r}_j(x)] -  \mathbb{E}_{x \sim \mathcal{X}}[\Vec{r}_i(x)] \cdot \mathbb{E}_{x \sim \mathcal{X}}[\Vec{r}_j(x)].
\]
The correlation variance is denoted as
\[
C_{ij} = \frac{\Sigma_{ij}}{\sqrt{\Sigma_{ii}\Sigma_{jj}}}.
\]

We rebalance the sources according to the eigenvector $\Vec{v}_{max} \in \mathbb{R}^{m+1}$ corresponding to the largest eigenvalue of $C$. In order to get reasonable weights, we first normalize $\Vec{v}_{max}$ such that the sum of the entries equals to one. Then we project $\Vec{v}_{max}$ to the weights simplex with minimal values $\frac{\epsilon}{m+1}$:
\begin{align*}
    w_{excess, j} &= \left(\Vec{v}_{max, j} - \frac{\epsilon}{m+1} \right)^+ \\
    \Vec{w}_{max} &= \frac{\epsilon}{m+1} + \frac{w_{excess}}{\|w_{excess}\|_1} \cdot (1 - \epsilon).
\end{align*}

This projection ensures that the weight on each source is at least $\epsilon$ portion of the value from equal weight, with the minimal ratio threshold $\epsilon \in [0,1]$. \\

Maximal eigenvalue method is recommended when the sources are relatively independent, and when the quality of the sources differ a lot. Intuitively, suppose two sources tends to agree with each other when they are not fitted well by the harmonizer, because there isn't intrinsic dependency between the sources, it is likely that the true label is given by the sources. Therefore, a maximal eigenvalue rebalancing scheme puts higher weights on the sources to encourage the harmonizer to fit to the unexplained examples. \\

\paragraph{Minimal Variance of Residual Correlation}
The same as in the maximal eigenvalue method, we consider the correlation matrix $C$ between the prediction residual $\Vec{r}_j$'s. Instead of finding the maximal eigenvalue of $C$, we consider solving the following minimal variance problem:
\[
\min_v v^T C v, \quad \mbox{s.t.} \ \mathbf{1}^T v = 1.
\]

This problem admits the closed form solution of
\[
v_{min} = \frac{C^{-1} \mathbf{1}}{\mathbf{1}^T C^{-1} \mathbf{1}}.
\]

Again, we project $\Vec{v}_{min}$ to the weights simplex with minimal values $\frac{\epsilon}{m+1}$:
\begin{align*}
    w_{excess, j} &= \left(\Vec{v}_{min, j} - \frac{\epsilon}{m+1} \right)^+ \\
    \Vec{w}_{min} &= \frac{\epsilon}{m+1} + \frac{w_{excess}}{\|w_{excess}\|_1} \cdot (1 - \epsilon),
\end{align*}
which ensures that the weight on each source is at least $\epsilon$ portion of the value from equal weight, with the minimal ratio threshold $\epsilon \in [0,1]$. \\

The minimal variance method is a classical portfolio rebalancing strategy in financial mathematics. The intuition behind the algorithm is minimizing the risk by diversification. This rebalancing scheme is useful when there are intrinsic dependency between the sources. Suppose two sources are duplicates and always tend to give the same label, their residuals should also be highly correlated. Minimal variance optimization automatically avoid putting too much weights on the duplicating sources. \\

While the equal weight method typically delivers good results in the simplest way, the other two rebalancing schemes are designed to address the specific concern such as source dependency and quality. For best performance, we recommend checking against the labels on a validation set if available.

\bigskip

\section{Training details}
\label{sect-train}
We explored different configurations of Pareto optimal learning below:
\begin{itemize}
    \item Harmonizer model: we experiment 1. BERT \cite{DBLP:journals/corr/abs-1810-04805} (PubMedBERT \cite{pubmedbert} for biomedical datasets CDR and ChemProt), 2. multi-layer perceptron (MLP), 3. Logistic regression (LR). The last two are built on top of the last layer embedding of the corresponding BERT model.
    \item Pareto loss scalarizer: we experiment all four loss scalarization functions as defined in Section \ref{sect-pol}, namely linear, quadratic, Euclidean norm, and Chebyshevy scalarization. 
    \item Optimizer: We use AdamW \cite{loshchilov2017decoupled} optimizer with learning rate $[10^{-4}, 10^{-5}, 10^{-6}]$, weight decay $[10^{-4}, 10^{-5}]$, batch size 16. All hyperparameters are optimized on held out dev set.
    \item Computation: We trained on Azure Standard NC12s v3 with 1 Nvidia V100 GPU.
\end{itemize}

\bigskip
\bigskip

\section{LLM Prompting Details}
\label{sect-prompt}
In this section we will describe the details of the prompts used to query the LLMs. 

\subsection{Out-of-the-box prompt}

\begin{itemize}
    \item Setting: describe the role of the LLM in the task, and the overall problem setting.
    \item Background: necessary background knowledge for domain specific tasks, including information from annotation guidelines for human annotators. 
    \item Data structure: for relation extraction tasks, explain the definition of each entity.
    \item Desired output: describe the list of the desired output. For each of the categories, provide explanation and optionally some examples.
    \item Chain of thought (CoT): instruction to encourage the LLM to think step-by-step, articulate point-by-point, and give the response in the desired structure.
    \item Confidence: ask the model to state ``unsure'' if it is not confident about the answer.
    \item Example: state the example and ask the model to perform the task.
\end{itemize}

Each prompt for out-of-the-box (zero-shot) prediction contains:
\begin{itemize}
    \item A problem setting part that depends on the specific dataset. 
    \item A response regularization part that encourages chain-of-thought (CoT) and confidence check, and specifies proper response format.
    \item A task instance part that contains the input instance and restates the task to perform.
\end{itemize}

\paragraph{Problem setting prompt}
\begin{itemize}
    \item CDR: ``You are an intelligent assistant to extract chemical-disease relations from academic literature. Your job is to determine if in the given piece of text, the drug (entity 1) induces the disease (entity 2) or not. Negative means the drug does NOT induce the disease. Positive means the drug induces the disease. Please use your judgement to the best of your knowledge. Your answer should be classified into the following categories: [Negative, Positive]. '' \\

    \item ChemProt: ``You are an intelligent assistant to extract chemical-protein interaction from academic literature. Your task is to identify the chemical-protein interactions (CHEMPROT) between entity 2: Chemical Entities Mentions (CEMs) and entity 1: Gene and Protein Related Objects (named as GPRO in the instruction below) in the given piece of text. In brief, the chemical-protein interactions include direct interactions (when a physical contact exits between a CEM and a GPRO, in most cases this GPRO being a protein or protein family and alters its function/activity) as well as indirect regulatory interactions between CEMs and GPROs (including genes, gene products (proteins, RNA), DNA/protein sequence elements and protein families, domains and complexes) that alter either the function or the quantity of the GPRO. The guidelines below provide curation rules to evaluate if the given sentence contains a description of a chemical-protein interaction; in particular, if sufficient detail/evidence is provided for comentioned CEMs and GPROs. Additionally, it describes curation rules and definitions to assign each identified chemical-protein interaction to any of the 10 classes, with detailed description listed below: 
    \begin{itemize}
        \item[] 0. Part of:  CEM that are structurally related to a GPRO: e.g. specific amino acid residues of a protein. 
        \item[]            1. Regulator: CEM that clearly regulates a GPRO, but for which there is no further information on whether the regulation is direct or indirect. 
        \item[]            2. Upregulator: CEM that increments a GPRO signal, without any insight on the mechanism. 
        \item[]            3. Downregulator: CEM that decreases a GPRO signal, without any insight on the mechanism. 
        \item[]            4. Agonist: CEM that binds to a receptor and alters the receptor state resulting in a biological response. 
        \item[]            5. Antagonist: CEM that reduces the action of another CEM, generally an agonist. Many antagonists act at the same receptor macromolecule as the agonist. 
        \item[]            6. Modulator: CEM that acts as allosteric modulator, compound that increases or decreases the action of an (primary or orthosteric) agonist or antagonist by combining with a distinct (allosteric or allotropic) site on the receptor macromolecule. 
        \item[]            7. Cofactor: CEM that is required for a protein's biological activity to happen. 
        \item[]            8. Substrate/Product: CEM that is both, substrate and product of enzymatic reaction. 
        \item[]            9. NOT: This class should be used to define the NEGATIVE occurrence of a chemical-protein interaction, without providing any further information on the specific negative CHEMPROT class or class.
    \end{itemize}
    Please identity the CHEMPROT interaction to the best of your knowledge. Your answer should be classified into the following categories: [Part of, Regulator, Upregulator, Downregulator, Agonist, Antagonist, Modulator, Cofactor, Substrate/Product, NOT]. '' \\

\item SemEval: ``You are an intelligent assistant to help recognize semantic relations between pairs of nomimals. For example, tea and ginseng are in an ENTITY-ORIGIN relation in "The cup contained tea from dried ginseng.". You will be given a piece of text, and Entity 1 and Entity 2 in the text for you to classify their semantic relation. The semantic relations are in the format of "entity1-entity2". The complete semantic relation inventory is given below: 
\begin{itemize}
    \item[] 0. Cause-Effect: An event or object (entity 1) leads to an effect (entity 2). Example: those cancers (entity 2) were caused by radiation exposures (entity 1)
            \item[]       1. Component-Whole: An object (entity 1) is a component of a larger whole (entity 2). Example: my apartment (entity 2) has a large kitchen (entity 1)
            \item[]       2. Content-Container: An object (entity 1) is physically stored in a delineated area of space (entity 2). Example: a bottle (entity 2) full of honey (entity 1) was weighed
            \item[]       3. Entity-Destination: An entity (entity 1) is moving towards a destination (entity 2). Example: the boy (entity 1) went to bed (entity 2)
            \item[]       4. Entity-Origin: An entity (entity 1) is coming or is derived from an origin (entity 2) (e.g., position or material). Example: letters (entity 1) from foreign countries (entity 2)
            \item[]       5. Instrument-Agency: An agent (entity 2) uses an instrument (entity 1). Example: phone (entity 1) operator (entity 2)
            \item[]       6. Member-Collection: A member (entity 1) forms a nonfunctional part of a collection (entity 2). Example: there are many trees (entity 1) in the forest (entity 2)
            \item[]       7. Message-Topic: A message (entity 1), written or spoken, is about a topic (entity 2). Example: the lecture (entity 1) was about semantics (entity 2)
            \item[]       8. Product-Producer: A producer (entity 2) causes a product (entity 1) to exist. Example: a factory (entity 2) manufactures suits (entity 1)
\end{itemize}
Please determine the semantic relation between entity 1 and entity 2 in the given text to the best of your knowledge. Your answer should be classified into the following categories: [Cause-Effect, Component-Whole, Content-Container, Entity-Destination, Entity-Origin, Instrument-Agency, Member-Collection, Message-Topic, Product-Producer]. ''

\item SMS: ``You are an intelligent assistant to determine if a text message is spam or not spam (ham). Your answer should be classified into the following categories: [ham, spam]. ''
    
\end{itemize}

\paragraph{Response regularization prompt}
``You may think step by step, articulate point by point, or make conclusion from multiple evidences, but please always state the most likely label as your answer at the very begining of your response. You are encouraged to reflect on your response, but please keep in mind that a clear answer is always desired. Try to give a clear answer at your best guess even when you are not very sure, in which case any of your conserns or explanations should go after the most likely answer to the best of your knowledge. If you are very unsure about the answer and are not willing to explicitly state any label, please say 'unsure' at the very begining of your response. ''

\paragraph{Task instance prompt}
\begin{itemize}
    \item Classification (for SMS): 
    
    ``Please classify the following example into the most likely category: [TEXT] ''
    \item Relation extraction (for CDR, ChemProt, SemEval): 
    
    ``Please classify the following example into the most likely category: [TEXT] Entity 1 [ENTITY 1] Entity 2: [ENTITY 2] ''
\end{itemize}

\bigskip

The complete prompt for querying the LLM is
\begin{align*}
    & \textbf{Problem setting prompt} \\
    + & \textbf{Response regularization prompt} \\
    + & \textbf{Task instance prompt}
\end{align*}

\bigskip

\subsection{Dynamic prompting}
In dynamic prompting, we query another follow-up prompt after the LLM gives the initial out-of-the-box response. As this is an extension to our main experiments, we only implemented for the CDR relation extraction task. The follow-up prompts for the two dynamic prompting strategies are:

\paragraph{Dynamic self-verification}
``Are you sure about your previous answer? If not, please give a new answer. Otherwise, please restate your previous answer. ''

\paragraph{POLAR-assisted RAG}
``It is possible that the answer could be something else. Here are some evidences to help you figure out the right answer. 
\[
\text{InformationSourceRetrival}(x, \ \vec{\lambda}(x))
\]
Are you sure about your previous answer? If not, please give a new answer. Otherwise, please restate your previous answer. ''

\bigskip

$\text{InformationSourceRetrival}(x, \ \vec{\lambda}(x))$ contains evidences from all the information sources $\lambda_j(x) \neq 0$ that are triggered by the input instance $x$. Examples of evidence from the information sources are shown below. Note that each evidence will be provided only when the corresponding information source is triggered.

\begin{itemize}
    \item ``According to the Comparative Toxicogenomics Database, the relation between the given chemical-condition pair is listed, confirming the answer. ''

\item  ``According to the Comparative Toxicogenomics Database, the given chemical-condition pair "[ENTITY 1]-[ENTITY 2]" is listed that the chemical actually treats the condition, so the answer that [ENTITY 1] does not induce [ENTITY 2] is confirmed. ''

\item ``According to the Comparative Toxicogenomics Database, the given chemical-condition pair "[ENTITY 1]-[ENTITY 2]" is listed that the chemical is typically present with the condition, which may confirm the answer if [ENTITY 1] induces [ENTITY 2]. ''

\item ``Based on the expression [INDUCE PATTERN], it is likely that [ENTITY 1] induces [ENTITY 2]. ''

\item ``Based on the expression [NOT INDUCE PATTERN], it is not likely that [ENTITY 1] induces [ENTITY 2]. ''

\item ``Based on the expression [C TREATS D PATTERN], [ENTITY 1] actually treats [ENTITY 2]. , so it is not likely that [ENTITY 1] induces [ENTITY 2]. ''

\item ``Based on the expression [CLOSE MENTION PATTERN], [ENTITY 1] is closely mentioned with [ENTITY 2], so they should be closely related. ''

\item ``Based on the expression [DISEASE IMPROVE PATTERN], the disease [ENTITY 2] is actually improved, so it is not likely that [ENTITY 1] induces [ENTITY 2]. ''

\item ``Based on the expression [INITIAL CONDITION PATTERN], [ENTITY 2] is the initial condition of the patient(s), so it is not likely that [ENTITY 1] induces [ENTITY 2]. ''

\item ``Based on the expression [UNCERTAIN PATTERN], it is uncertain that [ENTITY 1] induces [ENTITY 2]. ''

\item ``Based on the expression [INDUCED BY OTHER PATTERN], [ENTITY 2] is induced by other factors, so it is not likely that [ENTITY 1] induces [ENTITY 2]. ''

\item ``[ENTITY 1] and [ENTITY 2] are not closely mentioned in the text, so it is not likely that [ENTITY 1] induces [ENTITY 2]. ''

\item ``According to phrases like [WEAK EXPRESSION], there is no strong signal that [ENTITY 1] induces [ENTITY 2]. ''

\item ``According to the text, another chemical is mentioned closer to [ENTITY 2] than [ENTITY 1], so it is not likely that [ENTITY 1] induces [ENTITY 2]. ''

\item ``According to the text, another disease is mentioned closer to [ENTITY 1] than [ENTITY 2], so it is not likely that [ENTITY 1] induces [ENTITY 2]. ''

\end{itemize}

\end{document}